%% file: main.tex
\documentclass{article}
\usepackage{spconf,graphicx,hyperref}
\usepackage{amsmath,amssymb,amsfonts}
\usepackage{algorithm}
\usepackage{algpseudocode}
\usepackage{textcomp}
\usepackage{xcolor}
\usepackage{subfigure}
\usepackage{flushend}
\usepackage{multirow}
\usepackage{array}
\usepackage{lipsum}
\usepackage{capt-of}
\usepackage{cuted} 
\usepackage{booktabs}
\usepackage{graphicx}
\usepackage{subcaption}

\usepackage{siunitx}       
\usepackage{threeparttable}
\usepackage{caption}       
\title{Iterative Amortized Hierarchical VAE}
\name{S.W. Penninga\textsuperscript{\textdagger}, R.J.G. van Sloun
\thanks{This work was supported by the European Research Council (ERC) under the ERC starting grant nr. 101077368 (US-ACT).}}
\address{Eindhoven University of Technology, Department of Electrical Engineering \\
Eindhoven, The Netherlands\\
\textit{s.w.penninga@tue.nl}\textsuperscript{\textdagger}}

\begin{document}
\ninept
\maketitle

\input{abstract.tex}
\input{introduction.tex}

\begin{figure*}[ht]
    \centering
    \includegraphics[width=\linewidth]{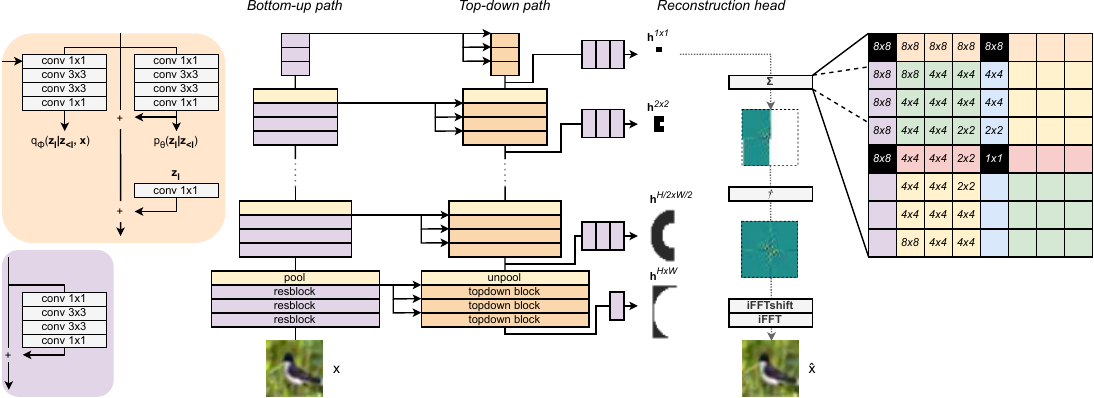}
    \caption{Schematic overview of the IA-HVAE architecture. The layers in the orange and purple blocks are shown on the left. A mapping between the model scales and spatial frequency bins is shown on the right for an example with $H\times W = 8\times 8$. Here the DC frequency is generated at the $1\times 1$ scale and the Nyquist frequencies at $8\times 8$. The missing values are filled in with the hermitian conjugate\textsuperscript{\textdagger} of the bins that share the same color to produce a symmetric spectrum and a purely real-valued image.}
    \label{fig:ia-hvae}
\end{figure*}

\input{methods.tex}

\input{experiments}
\input{conclusion}

\clearpage
\newpage
\bibliographystyle{IEEEbib}
\bibliography{references}

\end{document}

%% file: abstract.tex
\begin{abstract}
In this paper we propose the Iterative Amortized Hierarchical Variational Autoencoder (IA-HVAE), which expands on amortized inference with a hybrid scheme containing an initial amortized guess and iterative refinement with decoder gradients. We achieve this by creating a linearly separable decoder in a transform domain (e.g. Fourier space), enabling real-time applications with very high model depths. The architectural change leads to a 35x speed-up for iterative inference with respect to the traditional HVAE. We show that our hybrid approach outperforms fully amortized and fully iterative equivalents in accuracy and speed respectively. Moreover, the IA-HVAE shows improved reconstruction quality over a vanilla HVAE in inverse problems such as deblurring and denoising.
\end{abstract}
\begin{keywords}
Hierarchical VAE, Amortized inference, Iterative inference, Posterior inference, Inverse problems
\end{keywords}

%% file: introduction.tex
\section{Introduction}
\label{Introduction}
Inverse problems are a set of tasks in which the goal is to reconstruct a signal from degraded, incomplete, or noisy observations. Recently, interest has been sparked in addressing these tasks with deep generative models, since they can learn realistic high-dimensional priors and handle multi-modality. 

A popular initial approach was to search for a plausible solution in the range of a Generative Adversarial Network (GAN) via gradient descent \cite{Bora2017CompressedModels}. Later, this was surpassed by the current state-of-the-art Guided Diffusion \cite{Chung2025DiffusionProblems} in terms of reconstruction quality. This method leverages pre-trained diffusion models and iteratively guides the diffusion process to the observations to achieve conditional sampling. While this has shown promising results in a wide range of domains \cite{Stevens2025RemovingModels, Daras2024WarpedModels, Mardani2023AModels}, its application is bottle-necked by computational cost and the iterative backbone that cannot be parallelized. 

We propose to instead use the Hierarchical Variational AutoEncoder (HVAE) \cite{Snderby2016LadderAutoencoders} architecture for tasks where inference speed is crucial. This architecture has the potential to accelerate inference by initializing or replacing parts of the iterative process with a one-shot estimate for quicker convergence.

Although Variational AutoEncoders (VAEs) can perform rapid amortized inference, they have been largely overlooked for inverse problems as they are limited by an encoder that requires task-specific retraining which introduces an optimization bias. Next to this, they suffer from the \textit{amortization gap}, the error between the amortized posterior and the optimal posterior estimate \cite{Krishnan2018OnData, Cremer2018InferenceAutoencoders}. While VAEs are theoretically capable of performing exact posterior inference, the necessity of a variational family refrains them from doing so \cite{Hoffman2016ElboBound, Rosca2018DistributionInference, Wang2021PosteriorNon-identifiability}.
This has sparked interest in hybrid schemes that combine amortization of the data distribution with iterative refinement for cheap and accurate inference, balancing speed and precision \cite{Marino2018IterativeInference, Vafaii2025Brain-likeInference}. 
The HVAE architecture should not be confused with these hybrid models. Instead of incorporating iterative inference, it's amortized inference process is unrolled into multiple conditional steps, which we will call \textit{stacked inference}. Combining iterative optimization and stacked inference poses a challenge and requires architectural modifications to the HVAE as the computational cost of iterative optimization grows quadratically with the number of stochastic layers in the model. The performance of a HVAE relies heavily on having a large stochastic depth \cite{Child2021VeryImages}.
 Just like the diffusion process, the stacked inference of the HVAE is autoregressive and cannot be parallelized, leading to increased computational cost and latency. However, the improved performance over the traditional VAE has led to current state-of-the-art models for density estimation \cite{Kuzina2024HierarchicalVampPrior, Apostolopoulou2022DeepInference, Child2021VeryImages, Vahdat2020NVAE:Autoencoder}, outperforming even auto-regressive decoders \cite{Child2019GeneratingTransformers} and diffusion models \cite{Ho2020DenoisingModels, Kingma2021VariationalModels}, validating their potential for inference tasks and solving inverse problems.

In this paper we provide the first example of an Iterative Amortized HVAE (IA-HVAE), combining stacked inference of the HVAE with decoder gradient-based optimization. 
Inspired by latent aggregation \cite{Kuzina2024HierarchicalVampPrior}, we propose to introduce a HVAE decoder with a linear separation in outputs, giving every latent layer access to the gradient of their contribution to the reconstruction without the need to evaluate the rest of the hierarchy. 
We will show that this leads to decreased computational cost and faster iterative optimization. The new architecture outperforms amortized inference in terms of accuracy, and fully iterative inference in terms of latency. 
Next to this, we provide examples of its use in inverse problems, such as deblurring and denoising.

%% file: methods.tex
\section{Methods}
\label{sec:methods}
The goal of a VAE is to learn a joint distribution $p_\theta (x, z)$, usually factorized to $p_\theta (x|z)p_\theta(z)$ where $x$ denotes a signal in the input domain and $z \in \mathbf{Z}$ a set of latent variables. They are trained with the Evidence Lower Bound (ELBO): 
\begin{equation}
    \log p_\theta(x) \geq \mathbb{E}_{q_\phi(z|x)} \left[\log p_\theta (x|z)  \right] - D_{KL}[q_\phi(z|x) || p_\theta(z)],
    \label{eq:elbo}
\end{equation}
in which $\theta$ denotes the set of generative model parameters and $\phi$ the set of inference model parameters. $D_{KL}$ is the Kullback-Leibler divergence and for $p(z)$ we choose the Gaussian distribution.

\subsection{Iterative Amortized Hierarchical VAE}
We start with the HVAE architecture of the VDVAE \cite{Child2021VeryImages} that follows the conditioning structure of the LadderVAE \cite{Snderby2016LadderAutoencoders} in which the prior and approximate posterior are factorized in the same order:
\begin{equation}
p_\theta (\mathbf{z}) = p_{\theta_0} (\mathbf{z}_0)p_{\theta_1}(\mathbf{z}_1|\mathbf{z}_0) ... p_{\theta_L}(\mathbf{z}_L | \mathbf{z}_{<L}), 
\label{eq:prior}
\end{equation}
\begin{equation}
q_\phi(\mathbf{z}|\mathbf{x}) = q_{\phi_0}(\mathbf{z}_0|\mathbf{x})q_{\phi_1}(\mathbf{z}_1|\mathbf{z}_0,\mathbf{x}) ... q_{\phi_L}(\mathbf{z}_L|\mathbf{z}_{<L},\mathbf{x}).
\label{eq:posterior}
\end{equation}
Here, subscript $l$ indexes the stochastic latent layer and its accompanying network modules, with $l \in [0,L]$. 
Because of the similarity between the HVAE and diffusion models \cite{Luo2022UnderstandingPerspective}, we emphasize that this entire paper can be understood through the lens of diffusion models.
In this analogy, the weights of all $\theta_l$ are shared, $\mathbf{z}_l = x_t$ and $\phi_l$ is replaced with a Gaussian noising process $\forall l$.

The reconstruction $p_\theta(\mathbf{x}|\mathbf{z})$ is traditionally done through a neural network, acting as a single non-linear function on all the latent vectors.
Iterative posterior optimization through decoder gradients becomes intractable for deeper networks, since it requires the evaluation of all subsequent layers:
\begin{equation}
    \nabla_{\mathbf{z}_l}p_\theta(\mathbf{x}|\mathbf{z}) = \nabla_{\mathbf{z}_L}p_\theta(\mathbf{x}|\mathbf{z}_L) +\sum_{l'=l}^L\nabla_{\mathbf{z}_{l'}}p_\theta(\mathbf{z}_{l'}|\mathbf{z}_{<l'}).
    \label{eq:grad_pz}
\end{equation}
We propose to instead reconstruct $\mathbf{x}$ as a linear combination $\mathbf{B}$ of vectors $\mathbf{h}$ in a discrete set $\mathbf{H}$ through:

\begin{equation}
    \mathbf{x}(\mathbf{z}) = \sum_{\mathbf{h}\in\mathbf{H}}\mathbf{B}\mathbf{h}(\mathbf{z}).
\end{equation}
Now, the latent space is partitioned through a family of subsets $\{Z_\mathbf{h}\}_{\mathbf{h}\in\mathbf{H}}$ with $\bigcup_{\mathbf{h}\in\mathbf{H}} Z_\mathbf{h} = \mathbf{Z}$ and $Z_\mathbf{h} \bigcap Z_{\mathbf{h}'} = \emptyset$ $(\forall \mathbf{h}\neq \mathbf{h}')$ such that every $\mathbf{h}$ is created from a separate $Z_\mathbf{h}$, containing a unique [$\mathbf{z}_0, ..., \mathbf{z}_M$]. 
Instead of using $\nabla_{z_l}p_\theta(\mathbf{x}|\mathbf{z})$ we can now optimize a layer through:
\begin{equation}
    \nabla_{\mathbf{z}_m}p_\theta(\mathbf{h}|\mathbf{z}) = \nabla_{\mathbf{z}_m}p_\theta(\mathbf{h}|\mathbf{z}_M)+\sum^M_{m'=m}\nabla_{\mathbf{z}_{m'}}p_\theta(\mathbf{z}_{m'}|\mathbf{z}_{<m'}).
    \label{eq:grad_hz}
\end{equation}

\subsection{Implementation details}
\label{sec:implementation details}
The realization of $\mathbf{H}$ is flexible and can be matched with the inference task. In this paper we focus on the task of 2D image generation. In this domain the latent variables $\mathbf{z}$ do not necessarily share a single spatial resolution and are usually distributed over several network scales $s$ in order to capture both global and local structures. For example, the first latent vector $\mathbf{z}_0^s$ is often of scale $1\times 1$ and represents the DC frequency. The final latent $\mathbf{z}_L$ matches the data resolution $H\times W$ and is able to represent anything from DC to high-frequency noise \cite{Hazami2022Efficientvdvae:More}. This implicitly creates an ordering where the lowest frequency is created first, and higher order frequencies are conditionally generated through $p_\theta(\mathbf{z}^s|\mathbf{z}^{s-1})$. In this example implementation we copy this ordering for the decomposition, use the FFT as $\mathbf{B}$, and generate images in the frequency domain such that $\mathbf{H} = \{\mathbf{h}^{1\times1}, \mathbf{h}^{2\times2}, ..., \mathbf{h}^{H\times W}\}$ where every $\mathbf{h}$ contains the spatial frequencies of the respective scale. For domains that operate with real-valued signals, we predict only the half-spectrum, and obtain the full spectrum using the hermitian conjugate $(\dagger)$. A schematic overview of the architecture for real-valued signals together with an example of this frequency mapping $\mathbf{H} \rightarrow \mathbf{\hat{x}}$ is shown in Fig. \ref{fig:ia-hvae}. 

\subsection{Iterative optimization}
\label{sec:iterative optimization}
Performing iterative optimization of $\mathbf{z}$ solely with decoder gradients can move the latent vector off manifold and create instabilities for subsequent layers in the conditioning chain. We sample the distribution and refine the latent vector with Maximum A Posteriori (MAP) estimation. Using the negative log-likelihood (NLL) of the sample with respect to the prior we obtain the following update rule for a single latent vector:
\begin{equation}
    \mathbf{z}_l^{n+1} = \mathbf{z}_l^n - \lambda\nabla_{\mathbf{z}_l^n}\left[ \log\mathcal{N}(\mathbf{z}_l^n;\mu_p,\sigma_p) + \beta\mathcal{L}(\mathbf{h}^{s_l},\hat{\mathbf{h}}^{s_l})\right].
\end{equation}
Here $\lambda$ is the step size of the update, $\mu_p$ and $\sigma_p$ parameterize the prior distribution, $\mathcal{L}$ denotes a loss function and $\beta$ is a weight resembling guidance strength of the reconstruction. During the inference process, we sequentially update all latent vectors of the hierarchy in a Top-Down ordering.
The complete inference process including an amortized pass is shown in Algorithm \ref{alg:iterative_inference}.
\begin{algorithm}[th]
\caption{Inference sequence for the Iterative Amortized HVAE}\label{alg:iterative_inference}
\begin{algorithmic}
\State \textbf{Require:} input $\textbf{x}$, encoder parameters $\boldsymbol{\phi} = [\phi_0, ..., \phi_L]$, decoder parameters $\boldsymbol{\theta} = [\theta_0, ..., \theta_L]$, linear transformation $\mathbf{B}$, number of iterations $N$, subset length $M$, guidance strength $\beta$, step size $\lambda$, reconstruction loss function $\mathcal{L}$.

\State $[\mathbf{h}^{1\times 1}, ..., \mathbf{h}^{H\times W}] \gets \mathbf{B}^{-1}\mathbf{x}$ \textcolor{gray}{\Comment{Linear decomposition targets}}
\State $\mathbf{\hat{h}}_0 \gets \mathbf{0}$ \textcolor{gray}{\Comment{Initialize zero context vector}}

\For{$l = 1:L$}
    \State $\mathbf{z}^1_l \gets q_{\phi_l}(\mathbf{\hat{h}}_{l-1},\mathbf{x})$ \textcolor{gray}{\Comment{Single amortized pass}}
    \State $\boldsymbol{\mu}_p, \boldsymbol{\sigma}_p \gets p_{\theta_l}(\mathbf{\hat{h}}_{l-1})$ \textcolor{gray}{\Comment{Prior $p_\theta(\boldsymbol{\mu}_p, \boldsymbol{\sigma}_p|\mathbf{\hat{h}_{l-1}})$ for every $l$}}
    \For{$n = 1:N$}
        \State $\mathbf{\hat{h}}_l \gets p_{\theta_l}(\mathbf{\hat{h}}_{l-1}, \mathbf{z}^n_l)$ \textcolor{gray}{\Comment{Add contribution of latent $\mathbf{z}^n_l$}}
        \For{$m = l : M$} \textcolor{gray}{\Comment{Complete scale with subset $Z_\mathbf{h}$ (\ref{eq:grad_hz})}}
        \State $\mathbf{\hat{h}}_l \gets  p_{\theta_m}(\mathbf{\hat{h}}_l, \mathbf{z}_m)$ 
        \EndFor
        \State $\mathbf{z}_l^{n+1} \gets \mathbf{z}_l^n - \lambda \nabla_{z_l^n}\big[\log{\mathcal{N}(\mathbf{z}_l^n;\boldsymbol{\mu}_p, \boldsymbol{\sigma}_p)} + \beta \mathcal{L}(\mathbf{h}^{s_l}, \mathbf{\hat{h}}_l) \big]$
    \EndFor
\EndFor
\State $\mathbf{\hat{x}} \gets \sum_\mathbf{h} \mathbf{B}\mathbf{h}$ \\
\Return $\mathbf{\hat{x}}, \mathbf{z}$
\Statex \hspace{\algorithmicindent}\footnotesize\textbf{Note:} superscript denotes iterative step number $n$ for $\textbf{z}$, and the scale of the current layer $s_l$ for $\textbf{h}$.

\end{algorithmic}
\end{algorithm}

%% file: experiments.tex
\section{Experiments \& Results}
\label{sec:exp}
First, in Section \ref{sec:inference_time} we compare the iterative inference time of the IA-HVAE (\ref{eq:grad_hz}) with the vanilla HVAE architecture (\ref{eq:grad_pz}). Then in Section \ref{sec:inference_quality} we compare the inference quality of the vanilla amortized HVAE with that of a fully iterative inference using the IA-HVAE and the hybrid inference of the IA-HVAE. Lastly, the HVAE and the hybrid IA-HVAE are compared on inverse problems in Section \ref{sec:inverse_problems}.
To test performance on both real-valued and complex signals, we use CIFAR10 \cite{Krizhevsky2009LearningImages} ($32\times32$ pixels) and fastMRI \cite{Zbontar2018FastMRI:MRI} (rescaled to $128\times128$ pixels) respectively.

\subsection{Experimental setup}
For both datasets we train a model on the standardized train split and report the results on the test split. For CIFAR10 and fastMRI we have 6 and 8 scales respectively to populate all spatial frequencies. For CIFAR10 the model produces 3 half-spectra (real and imaginary values for every RGB channel) and for fastMRI a whole spectrum, as described in Section \ref{sec:implementation details}. 
Across all experiments the residual blocks have 128 outer and 64 inner channels with the swish activation function. The Bottom-up path and Reconstruction head of Fig. \ref{fig:ia-hvae} have 3 residual blocks per scale. For every latent vector we choose a channel size of 1 to maximize Active Units \cite{Burda2016ImportanceAutoencoders} in the model. We empirically find that a high number of Active Units corresponds to more efficient iterative optimization, because the vectors reside in a more compressed domain. All models are trained with a pixel-wise MSE loss and subsequent iterative refinement is done with an L1 loss on $\mathbf{H}$ with $\lambda=0.001$ and $\beta=1$ for all layers.
All results are gathered on a single NVIDIA GeForce 2080 Ti GPU, coded in  Keras 3.10.0 \cite{Chollet2015Keras} with the Tensorflow 2.19.0 \cite{Abadi2016TensorFlow:Learning} backend and XLA compiler. The comparisons between inference methods are done on the same model with the same weights.

\subsection{Inference time}
\label{sec:inference_time}
\begin{figure}[th]
    \centering
    \includegraphics[width=\linewidth]{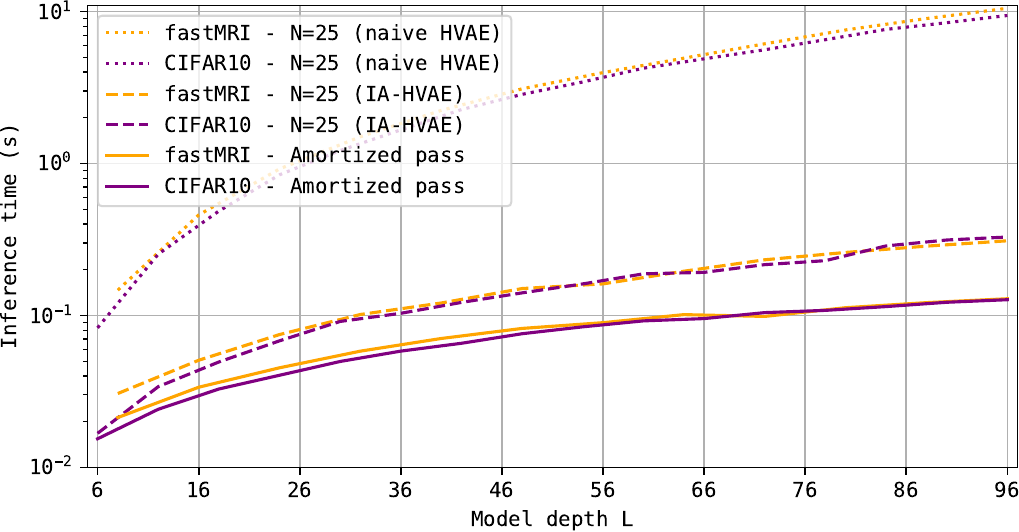}
    \caption{Inference time comparison between the vanilla HVAE architecture and the IA-HVAE architecture for different model depths.}
    \label{fig:time_results}
\end{figure}

A comparison of the inference time between the vanilla HVAE and the IA-HVAE is shown in Fig. \ref{fig:time_results}. 

\noindent For both datasets we show the execution time of a single amortized pass and that of 25 iterative refinement steps. The fact that there is no difference in timing between the datasets confirms that inference speed is limited solely by the model depth. 

The IA-HVAE shows favourable scaling with respect to depth, leading to a 35x speedup for deeper networks.
This performance gain would be even greater if the computational resources were saturated, as the vanilla HVAE architecture requires the entire model to be present in memory every iteration, leading to additional overhead.

\subsection{Inference quality}
\label{sec:inference_quality}
\begin{table*}[t]
\centering
\caption{Quantitative comparison of inference methods: amortized inference is the HVAE, iterative and hybrid inference the IA-HVAE.}
\label{tab:results}
\begin{threeparttable}
\begin{tabular}{@{} l c
    S[table-format=2.2] S[table-format=1.2] S[table-format=2.1] S[table-format=1.3]
    S[table-format=3.1] S[table-format=1.2] S[table-format=2.1] S[table-format=1.3] @{}}
\toprule
 & & \multicolumn{4}{c}{CIFAR10 (\textit{L=30})} & \multicolumn{4}{c}{fastMRI (\textit{L=40})} \\
\cmidrule(lr){3-6} \cmidrule(l){7-10}
Method & N
  & {MSE $\downarrow$} & {NLL (nats/dim) $\downarrow$} & {FID $\downarrow$} & {Time (s) $\downarrow$}
  & {MSE $\downarrow$} & {NLL (nats/dim) $\downarrow$} & {FID $\downarrow$} & {Time (s) $\downarrow$} \\
\midrule
\emph{Amortized inference} & 0
 & 18.27 & 0.86 & 31.6 & \textbf{0.051}
 & 161.2 & 0.69 & 47.1 & \textbf{0.081} \\
\addlinespace
\multicolumn{1}{@{}l}{\emph{Iterative inference}} & 5
 & 29.10\tnote{*} & \textbf{0.43}\tnote{*} & 77.6\tnote{*} & \underline{0.068}
 & 228.5\tnote{*} & \textbf{0.44}\tnote{*} & 98.2\tnote{*} & \underline{0.093} \\
 & 10
 & 24.91\tnote{*} & \underline{0.65}\tnote{*} & 55.4\tnote{*} & 0.074
 & 192.4\tnote{*} & \underline{0.48}\tnote{*} & 87.3\tnote{*} & 0.102 \\
 & 20
 & 22.75 & 0.71 & 37.0 & 0.095
 & 155.6 & 0.60 & 46.5 & 0.131 \\
 & 25
 & 20.54 & 0.73 & 34.5 & 0.103
 & 152.1 & 0.61 & 46.2 & 0.142 \\
 & 50
 & 18.01 & 0.78 & 31.0 & 0.193
 & 150.5 & 0.61 & 46.2 & 0.266 \\
\addlinespace
\multicolumn{1}{@{}l}{\emph{Hybrid inference}} & 5
 & 18.09 & 0.84 & 31.2 & 0.130
 & 158.2 & 0.65 & 47.0 & 0.162 \\
 & 10
 & 18.02 & 0.83 & 31.0 & 0.134
 & 153.3 & 0.62 & 46.4 & 0.169 \\
 & 20
 & 17.91 & 0.82 & 30.9 & 0.149
 & 149.6 & 0.61 & 46.1 & 0.186 \\
 & 25
 & \underline{17.86} & 0.80 & \underline{30.8} & 0.156
 & \underline{148.2} & 0.60 & \underline{45.9} & 0.192 \\
 & 50
 & \textbf{17.84} & 0.80 & \textbf{30.8} & 0.241
 & \textbf{145.0} & 0.59 & \textbf{45.5} & 0.293 \\
\bottomrule
\end{tabular}

\begin{tablenotes}
\footnotesize
\item Best performance shown in \textbf{bold}; second-best shown \underline{underlined}. Note that VAEs optimize for likelihood in training and not perceptual sharpness.
\item[*] A high MSE or FID with low NLL shows that iterative inference has not yet converged and the model is sampling close to the prior.
\end{tablenotes}
\end{threeparttable}
\end{table*}

\begin{figure}[ht] 
  \centering
  \begin{subfigure}
    \centering
    \includegraphics[width=\linewidth]{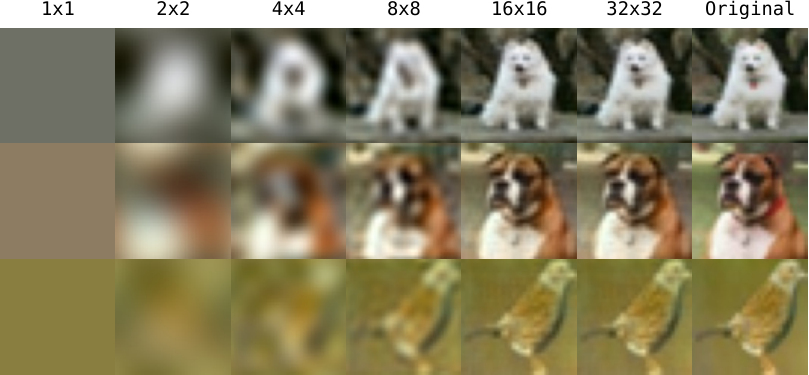}
    \label{fig:cifar10images}
  \end{subfigure}
  \begin{subfigure}
    \centering
    \includegraphics[width=\linewidth]{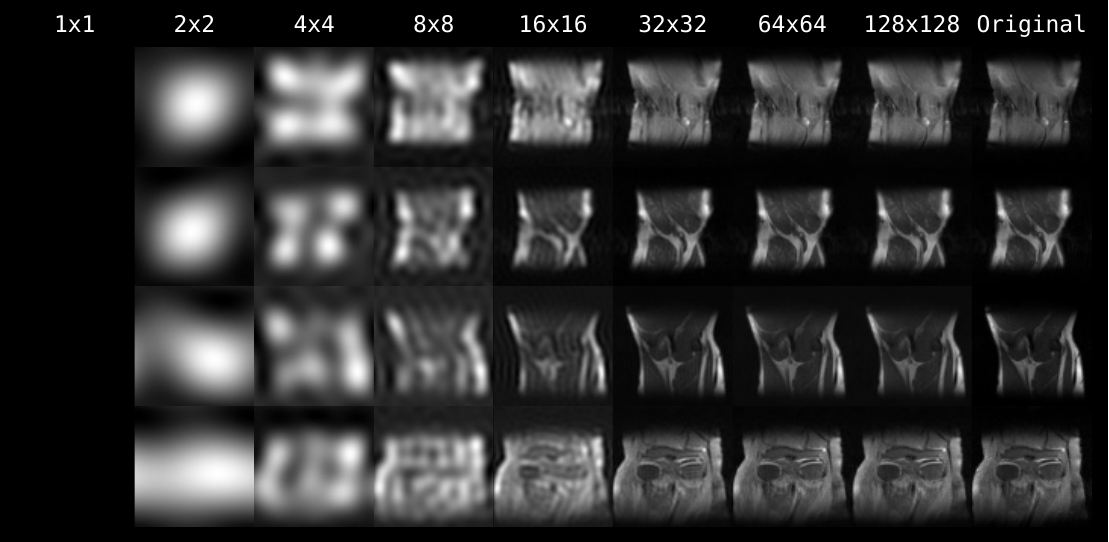}
    \label{fig:fastmriimages}
  \end{subfigure}
  \caption{Visualization of the IA-HVAE (\textit{N=25}) output at multiple scales for CIFAR10 and fastMRI. Best viewed zoomed in.}
  \label{fig:imageresults}
\end{figure}

To test inference quality we train a model for both datasets with 5 stochastic layers per scale (L=30 \& L=40).
A visualization of the IA-HVAE output is shown in Fig. \ref{fig:imageresults} at every scale, showing the frequency-wise conditioning in the generation process.

A quantitative comparison is shown in Table \ref{tab:results}. Note that the fastMRI dataset distribution is far more structured and less complex than CIFAR10. For this reason, iterative inference requires fewer iterations to converge. Nevertheless, for both datasets the hybrid approach always outperforms the amortized method in inference quality and can equate the quality of iterative inference in less time.

\subsection{Inverse problems}
\label{sec:inverse_problems}
\begin{figure}[ht]
    \centering
    \includegraphics[width=\linewidth]{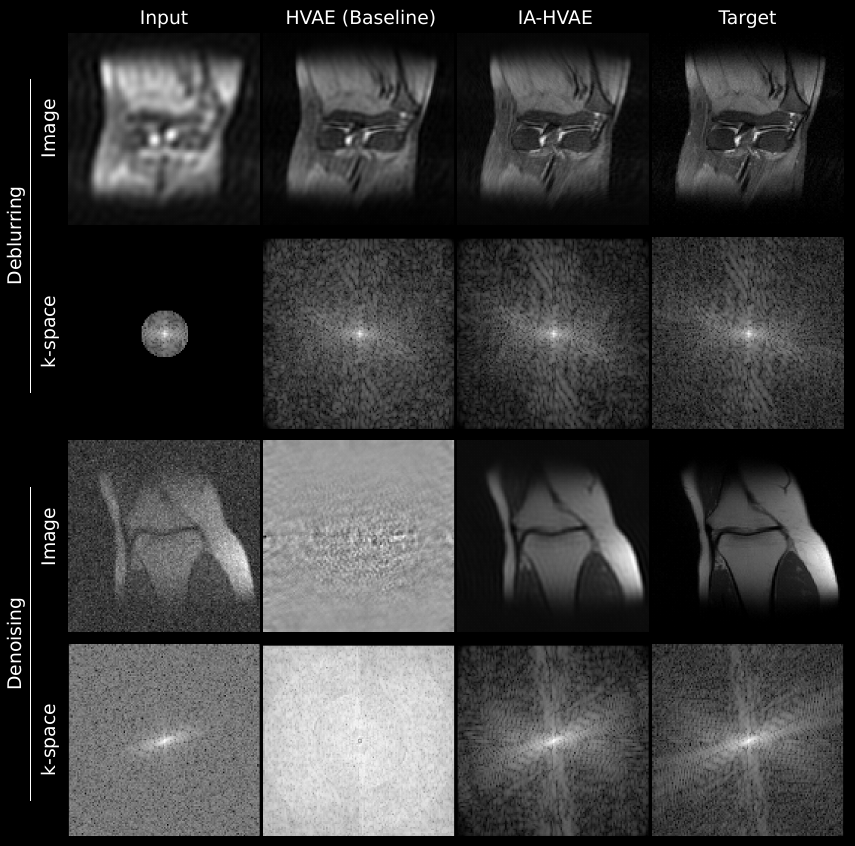}
    \caption{Qualitative evaluation of the IA-HVAE (\textit{N=25}) and vanilla HVAE (baseline) for deblurring and denoising experiments.}
    \label{fig:inverse}
\end{figure}

In this section we qualitatively compare the IA-HVAE (\textit{N=25}) performance to the vanilla HVAE on two inverse problems for the fastMRI dataset.
First we test deblurring. We measure a subset of the spatial frequency bins in an image up to $8\times 8$ and set others to 0. Since in MRI k-space coefficients are measured directly, this can also be interpreted as a compressed sensing setup. With both models we perform inference up until the $8\times 8$ scale, after which we continue reconstruction with the prior.
Next, we test a \textit{denoising} problem, in which we add $\mathcal{N}(0,1)$ noise to the model input which is normalized to the zero mean and unit variance.
The reconstruction results for both experiments are shown in Fig. \ref{fig:inverse} together with log-magnitude k-space representations.

The IA-HVAE outperforms the vanilla HVAE in both experiments.
With better inference it provides a better reconstruction for the deblurring example. It is also more successful in handling noisy images, since it is able to move latent vectors back to the data manifold when an amortized posterior estimate fails. 

%% file: conclusion.tex
\section{Discussion and Conclusion}
\label{sec:disc_and_concl}
Our implementation of the IA-HVAE with a frequency separated $\mathbf{H}$ outperformed the vanilla HVAE in inference quality and has shown promising application in deblurring and denoising. Future work could investigate layer or scale specific finetuning of $N$ to further decrease inference time.
We expect tasks that benefit from hierarchical conditioning, such as sequential and separable acquisition tasks, to benefit greatly from a specialized IA-HVAE implementation. 

One such example is any measurement over time that involves low frequency motion. The current diffusion-based methods require many noising steps to accommodate this type of pixel-wise translation in between frames, whereas the IA-HVAE can permute the latent layers at lower scales to express this phenomenon directly.

An obvious downside to the proposed approach is that the signal needs to be linearly decomposable. While this is possible for most fully sampled signals by moving to the frequency domain, this representation of the data can be more complex and harder for the model to learn. 
For future work, we suggest to research a more generalizable linear decomposition of signals that is still aligned with the Top-Down hierarchical nature of the model. 

Other future work includes the incorporation of iterative optimization into the training process. Performing decoder-based refinement  before the backward pass and weight update of the encoder will likely tighten the ELBO of the model. 
We leave a quantitative comparison with other generative architectures for future research.

%% file: references.bib
@article{Mardani2023AModels,
    title = {{A variational perspective on solving inverse problems with diffusion models}},
    year = {2023},
    journal = {arXiv preprint arXiv:2305.04391},
    author = {Mardani, Morteza and Song, Jiaming and Kautz, Jan and Vahdat, Arash}
}

@article{Vafaii2025Brain-likeInference,
    title = {{Brain-like variational inference}},
    year = {2025},
    journal = {ArXiv},
    author = {Vafaii, Hadi and Galor, Dekel and Yates, Jacob L},
    pages = {arXiv–2410}
}

@inproceedings{Bora2017CompressedModels,
    title = {{Compressed Sensing using Generative Models}},
    year = {2017},
    booktitle = {Proceedings of the 34th International Conference on Machine Learning},
    author = {Bora, Ashish and Jalal, Ajil and Price, Eric and Dimakis, Alexandros G},
    editor = {Precup, Doina and Teh, Yee Whye},
    month = {9},
    pages = {537--546},
    series = {Proceedings of Machine Learning Research},
    volume = {70},
    publisher = {PMLR},
    url = {https://proceedings.mlr.press/v70/bora17a.html}
}

@inproceedings{Apostolopoulou2022DeepInference,
    title = {{Deep Attentive Variational Inference}},
    year = {2022},
    booktitle = {ICLR 2022 - 10th International Conference on Learning Representations},
    author = {Apostolopoulou, Ifigeneia and Char, Ian and Rosenfeld, Elan and Dubrawski, Artur}
}

@article{Ho2020DenoisingModels,
    title = {{Denoising diffusion probabilistic models}},
    year = {2020},
    journal = {Advances in neural information processing systems},
    author = {Ho, Jonathan and Jain, Ajay and Abbeel, Pieter},
    pages = {6840--6851},
    volume = {33}
}

@article{Chung2025DiffusionProblems,
    title = {{Diffusion models for inverse problems}},
    year = {2025},
    journal = {arXiv preprint arXiv:2508.01975},
    author = {Chung, Hyungjin and Kim, Jeongsol and Ye, Jong Chul}
}

@article{Rosca2018DistributionInference,
    title = {{Distribution matching in variational inference}},
    year = {2018},
    journal = {arXiv preprint arXiv:1802.06847},
    author = {Rosca, Mihaela and Lakshminarayanan, Balaji and Mohamed, Shakir}
}

@article{Hazami2022Efficientvdvae:More,
    title = {{Efficientvdvae: Less is more}},
    year = {2022},
    journal = {arXiv preprint arXiv:2203.13751},
    author = {Hazami, Louay and Mama, Rayhane and Thurairatnam, Ragavan}
}

@inproceedings{Hoffman2016ElboBound,
    title = {{Elbo surgery: yet another way to carve up the variational evidence lower bound}},
    year = {2016},
    booktitle = {Workshop in Advances in Approximate Bayesian Inference},
    author = {Hoffman, Matthew D and Johnson, Matthew J},
    publisher = {NIPS}
}

@article{Zbontar2018FastMRI:MRI,
    title = {{fastMRI: An Open Dataset and Benchmarks for Accelerated MRI}},
    year = {2018},
    journal = {CoRR},
    author = {Zbontar, Jure and Knoll, Florian and Sriram, Anuroop and Murrell, Tullie and Huang, Zhengnan and Muckley et al., Matthew J},
    volume = {abs/1811.08839},
    url = {http://arxiv.org/abs/1811.08839}
}

@article{Child2019GeneratingTransformers,
    title = {{Generating long sequences with sparse transformers}},
    year = {2019},
    journal = {arXiv preprint arXiv:1904.10509},
    author = {Child, Rewon and Gray, Scott and Radford, Alec and Sutskever, Ilya}
}

@article{Kuzina2024HierarchicalVampPrior,
    title = {{Hierarchical VAE with a Diffusion-based VampPrior}},
    year = {2024},
    journal = {Transactions on Machine Learning Research},
    author = {Kuzina, Anna and Tomczak, Jakub M},
    pages = {1--21},
    volume = {2024},
    publisher = {Journal of Machine Learning Research (JMLR)}
}

@inproceedings{Burda2016ImportanceAutoencoders,
    title = {{Importance weighted autoencoders}},
    year = {2016},
    booktitle = {4th International Conference on Learning Representations, ICLR 2016 - Conference Track Proceedings},
    author = {Burda, Yuri and Grosse, Roger and Salakhutdinov, Ruslan}
}

@inproceedings{Cremer2018InferenceAutoencoders,
    title = {{Inference Suboptimality in Variational Autoencoders}},
    year = {2018},
    booktitle = {Proceedings of the 35th International Conference on Machine Learning},
    author = {Cremer, Chris and Li, Xuechen and Duvenaud, David},
    editor = {Dy, Jennifer and Krause, Andreas},
    month = {7},
    pages = {1078--1086},
    series = {Proceedings of Machine Learning Research},
    volume = {80},
    publisher = {PMLR},
    url = {https://proceedings.mlr.press/v80/cremer18a.html}
}

@inproceedings{Marino2018IterativeInference,
    title = {{Iterative amortized inference}},
    year = {2018},
    booktitle = {35th International Conference on Machine Learning, ICML 2018},
    author = {Marino, Joseph and Yue, Yisong and Mandt, Stephan},
    volume = {8}
}

@misc{Chollet2015Keras,
    title = {{Keras}},
    year = {2015},
    author = {Chollet, François and {others}},
    howpublished = {https://keras.io}
}

@inproceedings{Snderby2016LadderAutoencoders,
    title = {{Ladder variational autoencoders}},
    year = {2016},
    booktitle = {Advances in Neural Information Processing Systems},
    author = {S{\o}nderby, Casper Kaae and Raiko, Tapani and Maal{\o}e, Lars and S{\o}nderby, Søren Kaae and Winther, Ole},
    volume = {0},
    issn = {10495258}
}

@article{Krizhevsky2009LearningImages,
    title = {{Learning Multiple Layers of Features from Tiny Images}},
    year = {2009},
    journal = {{\ldots} Science Department, University of Toronto, Tech. {\ldots}},
    author = {Krizhevsky, Alex},
    doi = {10.1.1.222.9220},
    issn = {1098-6596}
}

@inproceedings{Vahdat2020NVAE:Autoencoder,
    title = {{NVAE: A deep hierarchical variational autoencoder}},
    year = {2020},
    booktitle = {Advances in Neural Information Processing Systems},
    author = {Vahdat, Arash and Kautz, Jan},
    volume = {2020-December},
    issn = {10495258}
}

@inproceedings{Krishnan2018OnData,
    title = {{On the challenges of learning with inference networks on sparse, high-dimensional data}},
    year = {2018},
    booktitle = {Proceedings of the Twenty-First International Conference on Artificial Intelligence and Statistics},
    author = {Krishnan, Rahul and Liang, Dawen and Hoffman, Matthew},
    editor = {Storkey, Amos and Perez-Cruz, Fernando},
    month = {7},
    pages = {143--151},
    series = {Proceedings of Machine Learning Research},
    volume = {84},
    publisher = {PMLR},
    url = {https://proceedings.mlr.press/v84/krishnan18a.html}
}

@inproceedings{Wang2021PosteriorNon-identifiability,
    title = {{Posterior Collapse and Latent Variable Non-identifiability}},
    year = {2021},
    booktitle = {Advances in Neural Information Processing Systems},
    author = {Wang, Yixin and Blei, David M. and Cunningham, John P.},
    volume = {7},
    issn = {10495258}
}

@article{Stevens2025RemovingModels,
    title = {{Removing Structured Noise using Diffusion Models}},
    year = {2025},
    journal = {Transactions on Machine Learning Research},
    author = {Stevens, Tristan and van Gorp, Hans and Meral, Faik C and Shin, Junseob and Yu, Jason and Robert, Jean-luc and Sloun, Ruud Van},
    url = {https://openreview.net/forum?id=BvKYsaOVEn},
    issn = {2835-8856}
}

@inproceedings{Abadi2016TensorFlow:Learning,
    title = {{TensorFlow: A system for large-scale machine learning}},
    year = {2016},
    booktitle = {Proceedings of the 12th USENIX Symposium on Operating Systems Design and Implementation, OSDI 2016},
    author = {Abadi, Martin and Agarwal, Ashish and Barham, Paul and Brevdo, Eugene and Chen, Zhifeng and Citro et al., Craig}
}

@article{Luo2022UnderstandingPerspective,
    title = {{Understanding diffusion models: A unified perspective}},
    year = {2022},
    journal = {arXiv preprint arXiv:2208.11970},
    author = {Luo, Calvin}
}

@article{Kingma2021VariationalModels,
    title = {{Variational diffusion models}},
    year = {2021},
    journal = {Advances in neural information processing systems},
    author = {Kingma, Diederik and Salimans, Tim and Poole, Ben and Ho, Jonathan},
    pages = {21696--21707},
    volume = {34}
}

@inproceedings{Child2021VeryImages,
    title = {{Very Deep VAEs Generalize Autoregressive Models and Can Outperform Them on Images}},
    year = {2021},
    booktitle = {ICLR 2021 - 9th International Conference on Learning Representations},
    author = {Child, Rewon}
}

@inproceedings{Daras2024WarpedModels,
    title = {{Warped Diffusion: Solving Video Inverse Problems with Image Diffusion Models}},
    year = {2024},
    booktitle = {Advances in Neural Information Processing Systems},
    author = {Daras, Giannis and Nie, Weili and Kreis, Karsten and Dimakis, Alexandros G and Mardani, Morteza and Kovachki, Nikola B and Vahdat, Arash},
    editor = {Globerson, A and Mackey, L and Belgrave, D and Fan, A and Paquet, U and Tomczak, J and Zhang, C},
    pages = {101116--101143},
    volume = {37},
    publisher = {Curran Associates, Inc.},
    url = {https://proceedings.neurips.cc/paper_files/paper/2024/file/b736c4b0b38876c9249db9bd900c1a86-Paper-Conference.pdf}
}
